# Large Language Models for 3D IC Space Planning

Achieving Zero-Dead-Space Layouts with Post-Order Slicing Tree Encoding

HUNG-YING CHU*

Dept. of Computer Science & Engineering, Yuan Ze University

s1111521@mail.yzu.edu.tw

GUAN-WEI CHEN*

Dept. of Computer Science & Engineering, Yuan Ze University

s1111446@mail.yzu.edu.tw

SHAO-YU WEI*

Dept. of Computer Science & Engineering, Yuan Ze University

s1123332@mail.yzu.edu.tw

YU-CHENG LIN*

Dept. of Computer Science & Engineering, Yuan Ze University

linyu@saturn.yzu.edu.tw

Three-dimensional integrated circuits (3D ICs) have emerged as a promising solution to the scaling limits of two-dimensional designs, offering higher integration density, shorter interconnects, and improved performance. As design complexity increases, effective space planning becomes essential to reduce dead space and ensure layout quality. This study investigates the use of large language models (LLMs) for 3D IC space planning through a post-order slicing tree representation, which guarantees legal space plans while aiming to minimize dead space. Open-source LLMs were fine-tuned on large-scale synthetic datasets and further evaluated on MCNC-derived 3D benchmarks. Experimental results indicate that the proposed framework achieves a favorable balance between runtime efficiency, legality, and dead-space reduction, with zero-dead-space layouts obtained in a significant portion of test cases under practical runtime budgets. Beyond synthetic benchmarks, the method generalizes to MCNC cases such as ami33 and ami49, though larger and irregular instances remain challenging. The approach also shows potential for cross-domain applications, including logistics and 3D object placement, where spatial efficiency is critical. Overall, the results suggest that LLM-based space planning can serve as a data-driven complement to traditional electronic design automation (EDA) methods, providing new insights for scalable 3D layout generation.

# 1  INTRODUCTION

As semiconductor size approaches physical limits, traditional two-dimensional integrated circuits (2D ICs) are increasingly limited by area, delay, and power consumption. Moore's Law alone is no longer sufficient to maintain performance growth [1]. To address this challenge, three-dimensional integrated circuits (3D ICs) have emerged as a promising "Beyond Moore" solution [2]. By vertically stacking chip modules, 3D ICs can significantly shorten interconnect lengths, improve computing performance, and reduce power consumption, making them a key technology in the post-Moore era.

However, the development of 3D ICs has also brought about more complex layout planning problems. In addition to properly arranging modules to minimize dead space and wirelength, designers must also consider power distribution, thermal management, and through-silicon via (TSV) layout. This type of highly constrained combinatorial optimization problem is classified as NP-hard and has long been considered one of the core challenges of electronic design automation (EDA) [3]. Over the past few decades, many heuristic methods have been developed, such as simulated annealing (SA) based on slicing trees [4] and B* trees [5]. Although SA can escape local optima through random search, its running time and convergence cost grow rapidly with the number of modules, and the quality of the solution often fluctuates due to its inherent randomness.

In recent years, advances in artificial intelligence have brought new opportunities to this field. In particular, deep reinforcement learning (DRL) has shown strong potential in optimizing both area and wirelength simultaneously, surpassing traditional heuristics on multiple benchmark datasets [6,7,8]. At the same time, large language models (LLMs) have demonstrated excellent capabilities in sequence modeling and structure generation, providing a new paradigm for layout planning. Although only a few studies have explored this application, Lu and Yeh [9] successfully applied LLMs to 2D integrated circuit floorplanning, showing that data-driven LLMs can produce high-quality solutions with reduced dead space and greater stability than randomized algorithms.

Building on this emerging direction, this study proposes a fine-tuning framework that adapts LLMs specifically for 3D IC space planning. We fine-tune open-source LLMs on large-scale synthetic datasets and further extend evaluation to MCNC-derived 3D benchmarks (ami33, ami49, hp, xerox, apte), ensuring that the experiments are aligned with widely adopted design standards. Comparative experiments with SA under identical runtime budgets indicate that the proposed approach can achieve competitive runtime efficiency, lower dead-space ratios, and more stable solutions. While challenges remain for larger and irregular MCNC cases, the results confirm that LLM-based planning is a promising complement to traditional optimization in practical design contexts.

Contributions of this work include:
1. Proposing an LLM-based fine-tuning framework tailored for 3D IC space planning with post-order slicing tree encoding.
2. Demonstrating improved efficiency and reduced dead space compared to SA under the same runtime budgets.
3. Extending evaluation to MCNC-derived 3D benchmarks, bridging synthetic datasets with realistic industry-standard designs.
4. Showing adaptability of the framework to broader domains such as logistics and 3D object placement.

These contributions demonstrate the feasibility of LLM as a data-driven alternative to traditional optimization methods in electronic design automation (EDA), combining efficiency and versatility.

## 2 PROBLEM DESCRIPTION

In the early stages of integrated circuit design, space planning plays a critical role, with the objective of assigning appropriate spatial regions to each module as the foundation for subsequent physical design. Given a set of modules $P = \{p_0, p_1, p_2, ..., p_n\}$, where each module $p_i$ is represented as a rectangular cuboid with fixed dimensions $(w_i, h_i, d_i)$, the goal is to place these cuboids within a three-dimensional space so as to minimize dead space, as illustrated in Figure 1.

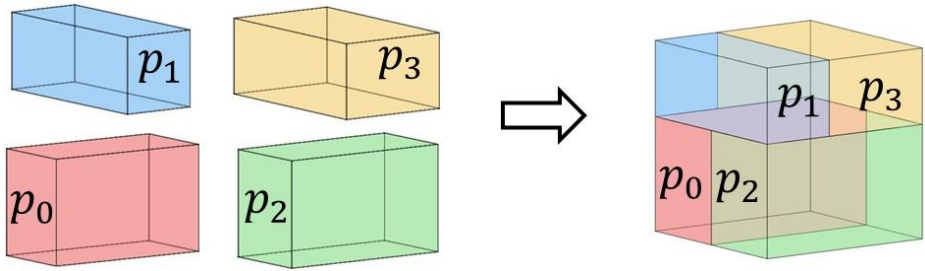

Figure 1. space planning problem with four modules

The most straightforward way to fully describe the solution to a floor plan is to record the coordinates of each module in three-dimensional space, including the reference position of its lower left corner (or lower front corner) and its width, height, and depth. While this representation uniquely identifies a solution, as the number of modules increases, storing all coordinates becomes too large for algorithms to operate on. As the number of modules increases, the dimensionality of the solution space rapidly expands, making it increasingly difficult to find the optimal solution.

To represent space planning results more efficiently, this study adopts the three-dimensional slicing tree (3D slicing tree) as the primary representation. A slicing tree recursively partitions the chip outline through cutting operations—horizontal (H), vertical (V), and depth (D)—until the space is divided into a set of non-overlapping subregions, each containing one module. With only a tree structure and a sequence of cut directions, a valid space plan can be fully reconstructed, thereby avoiding the complexity of explicitly recording the coordinates of every module, as illustrated in Figure 2.

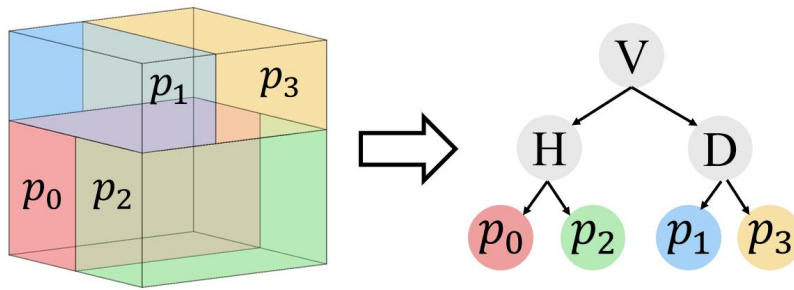

Figure 2. Turning a slicing space plan into a slicing tree

In this study, we leverage large language models (LLMs) to directly generate post-order representations of slice trees and use them to construct a three-dimensional space planning solution with the goal of minimizing dead zones. To formally describe dead zones, we employ a three-dimensional slice tree (3D slice tree) as the primary space planning representation. The slice tree recursively merges modules into larger regions through partitioning operations, ultimately forming a complete wafer layout. Each slice operation merges two modules (or sub-regions) into a new parent region, corresponding to three basic operations:

(1) Horizontal cut (H): along the X-axis, with parent dimensions:
$$W_{parent} = w_i + w_j;\ h_{parent} = max(h_i, h_j);\ d_{parent} = max(d_i, d_j)$$
(2) Vertical cut (V): along the Y-axis, with parent dimensions:
$$W_{parent} = max(w_i, w_j);\ h_{parent} = h_i + h_j;\ d_{parent} = max(d_i, d_j)$$
(3) Depth cut (D): along the Z-axis, with parent dimensions:
$$W_{parent} = max(w_i, w_j);\ h_{parent} = max(h_i, h_j);\ d_{parent} = d_i + d_j$$

Dead space is defined as the unused volume that arises when two modules or sub-regions are merged during the slicing tree construction process. For any two modules $p_i$ and $p_j$, the dead space is calculated as:
$$ds(p_i, p_j) = V_{parent} - V_i - V_j \quad (1)$$

where $V_i = w_i \times h_i \times d_i$, $V_j = w_j \times h_j \times d_j$ and the parent volume $V_{parent}$ depends on the cut direction:
$$\begin{cases} Horizontal\ cut: V_{parent} = (w_i + w_j) \times max(h_i, h_j) \times max(d_i, d_j) & (2) \\ Vertical\quad cut: V_{parent} = max(w_i, w_j) \times (h_i + h_j) \times max(d_i, d_j) & (3) \\ Depth\quad\ \ cut: V_{parent} = max(w_i, w_j) \times max(h_i, h_j) \times (d_i + d_j) & (4) \end{cases}$$

The total dead space for the entire space plan is the sum of dead spaces generated by all merging operations:
$$TotalDeadSpace = \sum_{k=1}^{n-1} dead\_space_k \quad (5)$$

Figure 3. illustrates the formation of dead space during the construction of a 3D slicing floorplan. In (a), three modules $p_0$, $p_1$, and $p_2$ are arranged according to a vertical–horizontal slicing tree, resulting in a non-optimal configuration. The corresponding slicing tree is shown in (b). In (c), modules $p_0$ and $p_1$ are merged by a horizontal cut, producing an unused region $DS_1$ (highlighted with a red dashed box) due to the height mismatch. In (d), the composite module $p_1'$ is further merged with $p_2$ by a depth cut, generating an additional unused region $DS_2$. These dead spaces arise whenever the bounding box of a merged region exceeds the total occupied volume of its child modules, and the total dead space of the floorplan is obtained by summing all such regions across the slicing process.

We can traverse the 3D slicing tree in post-order, calculate the dead space between two adjacent modules according to their spatial relationship (horizontal, vertical, or depth-wise adjacency), merge them into a composite module, and update the three-dimensional dimensions accordingly. The total dead space for the entire 3D space plan is computed as the sum of all individual dead space calculations throughout the merging process. The optimal 3D space plan is defined as the one that minimizes the total dead space while satisfying all geometric and placement constraints.

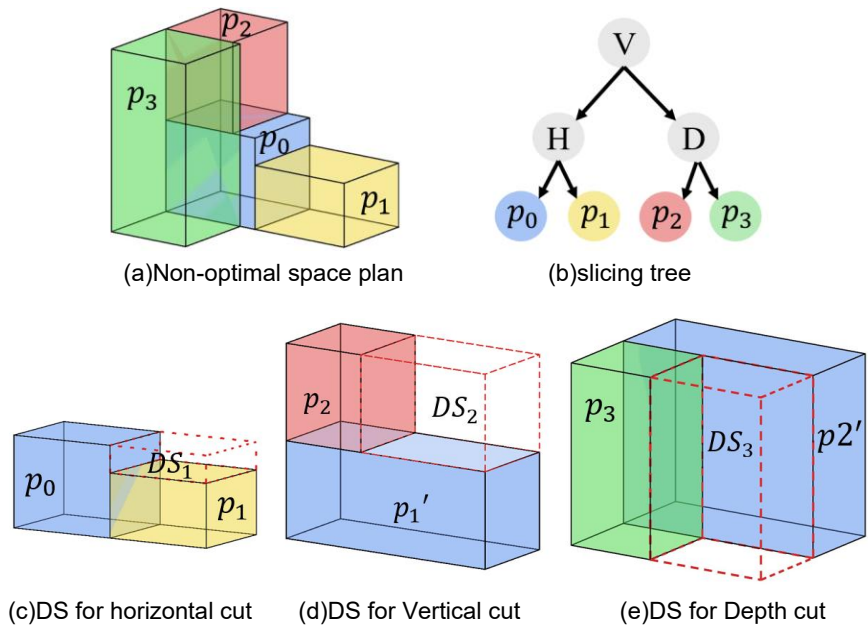

(a)Non-optimal space plan  (b)slicing tree

(c)DS for horizontal cut   (d)DS for Vertical cut   (e)DS for Depth cut

Figure 3. Dead space example in horizontal cut operation (DS: Dead Space).

## 3 METHODOLOGY

In this study, our methodology is divided into three main stages: the dataset generation stage, the model fine-tuning stage, and the inference and dead-space evaluation stage. The overall workflow is illustrated in Figure 4.

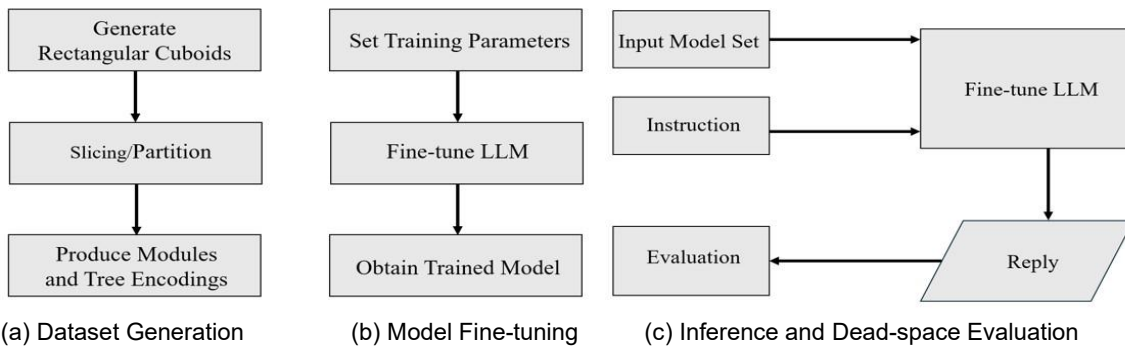

(a) Dataset Generation   (b) Model Fine-tuning   (c) Inference and Dead-space Evaluation

Figure 4. Three stage flow chart

### 3.1 Dataset Generation Stage

#### 3.1.1 Block Generation:

A cuboid representing the chip is first generated, with its height, width, and depth randomly sampled.

#### 3.1.2 Recursive Partitioning:

- At each step, one cuboid is randomly selected and split along one of three directions: horizontal (H), vertical (V), or depth (D).
- Each cut ensures that the cuboid is divided into two modules, with two dimensions preserved depending on the cut direction.

- The sub-block closer to the origin is consistently designated as the left child node.
- This process continues recursively until the specified number of modules is reached.

### 3.1.3 Module Generation and Tree Encoding:

The resulting slicing tree is encoded using post-order traversal. Each module is labeled as $p_0, p_1, p_2, \ldots, p_n$, while cutting operations are represented by H, V, D symbols. These tokens are sequentially inserted into the encoding, forming a complete post-order expression that includes both module and cut information, as illustrated in Figure 5.

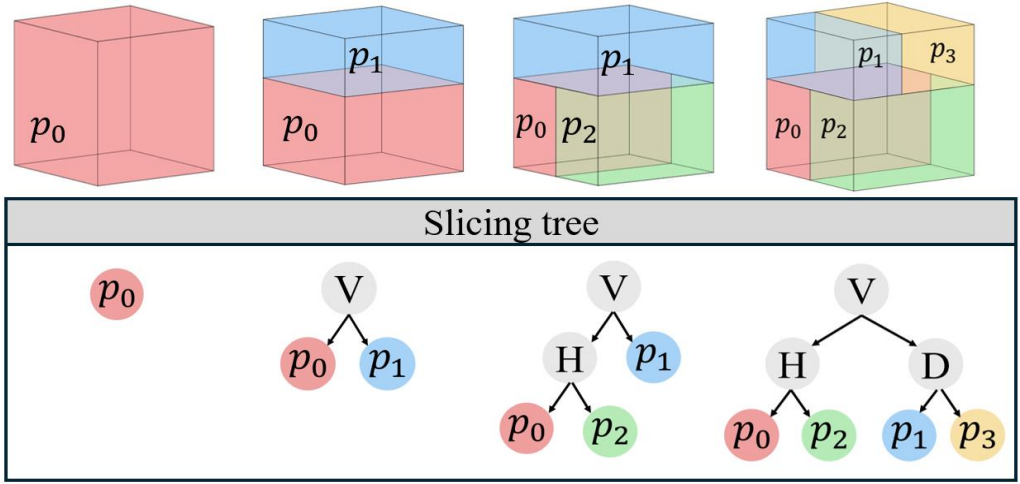

(a) Generate a cuboid    (b) Vertical cut    (c) Horizontal cut    (d) Depth cut

Figure 5. Illustration of 3D slicing operations and corresponding slicing tree representations.

### 3.1.4 MCNC-derived Instance Construction:

To evaluate the generalization ability of our framework on realistic footprints, we further derive three-dimensional instances from widely used MCNC floorplanning benchmarks, including ami33, ami49, hp, xerox, and apte. For each MCNC case, the original two-dimensional module widths and heights are preserved, while a depth value is assigned to every module through random sampling within the instance-level range $[d_{min}, d_{max}]$, where $d_{min}$ and $d_{max}$ are the minimum and maximum lateral dimensions observed in that case. The resulting modules are then encoded into post-order slicing trees using the same stack-based method described in Section 3.1.3. This procedure ensures that the generated 3D inputs retain the aspect characteristics of MCNC cases while introducing realistic variability along the depth dimension.

For larger benchmarks such as ami33 and ami49, directly training or inferring with 33 or 49 modules poses significant challenges due to the sequence length and the limited context window of current LLMs. To address this, we adopt a grouping strategy: ami33 is partitioned into 11 volumetrically balanced composite modules, while ami49 is divided into five groups of sizes (10, 10, 10, 10, 9). Each composite preserves the original footprint characteristics while reducing the effective number of modules, thereby aligning with the 8–16 module range used in synthetic datasets.

After grouping, we compute the bounding boxes of each composite module to obtain their effective dimensions. These composite modules are then passed to the fine-tuned LLM, which generates the corresponding post-order slicing trees. The reconstructed layouts are subsequently refined through a lightweight one-dimensional compaction applied sequentially along the X, Y, and Z axes, which eliminates residual gaps while preserving legality.

This preprocessing pipeline — grouping → bounding box calculation → LLM inference → compaction — enables tractable evaluation of large-scale MCNC benchmarks while maintaining fidelity to their original structural characteristics. It also provides a practical bridge between the synthetic training regime and real-world benchmarks, ensuring consistency across different experimental settings.

In this study, we adopt the post-order expression to encode slicing trees. Compared with pre-order, post-order provides a simpler structure: modules appear first in the sequence, followed by cut operations (H, V, D). This allows the slicing tree to be uniquely reconstructed using a stack-based method, avoiding parsing ambiguity. Post-order representation has also been widely adopted in prior works, ensuring consistency and comparability.

## 3.2 Model Fine-Tuning Stage

### 3.2.1 Parameter Configuration:
Before fine-tuning, we configure hyperparameters such as learning rate, batch size, number of epochs, and gradient accumulation steps. To reduce memory requirements, LoRA/QLoRA is applied.

### 3.2.2 Fine-Tuning Process:
Supervised fine-tuning is conducted using the input-output pairs generated in the dataset stage. The goal is to train the model to generate the correct post-order slicing tree expression given a set of modules.

### 3.2.3 Model Output:
The fine-tuned model outputs only module labels and cut operators (H/V/D), separated by semicolons, ensuring that each generated sequence corresponds uniquely to a valid slicing tree.

## 3.3 Inference and Dead-Space Evaluation Stage

### 3.3.1 Module Input and Post-Order Output:
During inference, a new set of modules is provided, and the fine-tuned LLM outputs the corresponding post-order slicing tree expression

### 3.3.2 Tree Reconstruction and space plan Generation:
The generated post-order expression is reconstructed via a stack-based algorithm to produce the 3D space plan. A lightweight one-dimensional compaction is then applied sequentially along X, Y, and Z to remove residual gaps while preserving legality.

### 3.3.3 Dead-Space Calculation and Evaluation:
- **Legality (L):** The percentage of test cases that yield valid 3D slicing trees.

- **Best Dead-Space Ratio (B):** For each problem, the solution with the minimum dead space among all candidates, defined as the ratio of total dead space to the volume of the root bounding box.
- **Global Dead-Space Ratio (G):** The average dead-space ratio across all problems and candidate solutions.
- **Across-Case Best Average (A):** The mean of the best dead-space ratios per problem.

These metrics are formally defined as:

$$L = \frac{N_{legal}}{N_{total}} \times 100\% \quad (6) \qquad B = \frac{total\_dead}{root\_volume} \quad (7)$$

$$G = \frac{\Sigma daed\_ratio}{N_{cases}} \quad (8) \qquad A = \frac{\Sigma best\_dead\_ratio\_root}{N_{questions}} \quad (9)$$

Where $N_{legal}$ is the number of valid slicing trees, $N_{total}$ is the total number of test cases, $total\_dead$ is the total dead space volume, $root\_volume$ is the final bounding box volume, $daed\_ratio$ is the dead-space ratio of each candidate solution, $N_{cases}$ is the number of all candidates, $best\_dead\_ratio\_root$ is the best dead-space ratio per test case, and $N_{questions}$ is the total number of test problems.

### 3.3.4 Comparison with Simulated Annealing:

Under identical time constraints, the space plans generated by the LLM are evaluated against two variants of simulated annealing (SA): SA-Fast, a configuration with limited runtime designed to emphasize computational efficiency, and SA-Quality, a quality-oriented baseline with a runtime of approximately 15 minutes. To ensure fairness, both the LLM and SA solutions are further refined with the same one-dimensional compaction procedure. These baselines serve as references to demonstrate the advantages of our method in both dead-space reduction and runtime efficiency.

*SA-Fast was implemented with k = 1 , α=0.70, batch size=125, $T_{min} = 1$;*
*SA-Quality was implemented with k = 2 , α=0.97, batch size=200, $T_{min} = 1e - 3$.*

## 3.4 Experimental Setup

This study employs the Unsloth [10] framework to accelerate training, combined with open-source LLMs for fine-tuning.

### 3.4.1 Dataset Specification:

- **Number of Modules:** For the synthetic dataset, we select 8 to 16 modules per problem instance. This range avoids trivial cases with too few modules and prevents excessive computational overhead with too many modules, while corresponding to typical sub-block sizes in practical chip partitioning. For the enlarged-cuboid dataset, we restrict the range to 9–11 modules in order to approximate MCNC benchmarks.
- **Dataset Size:** The training corpus consists of two parts: 120,000 synthetic slicing-tree instances (8–16 modules) and 200,000 enlarged-cuboid instances (9–11 modules) generated by scaling base cuboids to approximate MCNC dimensions. The latter exposes the model to larger geometries and more realistic aspect ratios.

- **Test Samples:** Each model is first evaluated on 100 unseen synthetic cases generated using the same procedure as the training data. To further validate generalization, we prepare 50 additional test cases derived from MCNC benchmarks, including ami33, ami49, hp, xerox, and apte. For each case, the original 2D dimensions are preserved, while depths are assigned to modules through random sampling within the instance-level range $[d_{min}, d_{max}]$. For ami33, 33 modules are grouped into 11 volumetrically balanced composites; for ami49, modules are partitioned into five groups of sizes (10, 10, 10, 10, 9).

### 3.4.2 Fine-Tuning Details:

We fine-tuned four open-source models of different scales: LLaMA3.1 (8B) [11], LLaMA3.2 (3B) [12], GPT-2 [13], and Phi-4 (13B) [14]. These models span from large to small sizes, enabling us to examine performance variations across scales and architectures. Among them, LLaMA3.1 (8B) achieved the best performance on 8-module problems and was selected as the primary model for comparison with SA.

In addition to the general training on 120k synthetic slicing-tree cases, we further fine-tuned a specialized LLaMA3.1 (8B) using the 200k enlarged-cuboid dataset (Section 3.4.1). This dataset was designed to approximate the scale of MCNC benchmarks and restricted to 9–11 modules, enabling the model to better adapt to MCNC-derived tests (see Section 3.4.1) without repeating grouping details.

All training was performed on a server equipped with an NVIDIA V100 GPU (32GB), using the Unsloth framework. The training time for each model was approximately 3 hours, except for Phi-4 (13B), which required around 17 hours due to its significantly larger model size and longer convergence time.

## 4 EXPERIMENTAL RESULT

On the 8-module benchmark, we first compared four LLMs of different scales: LLaMA3.1 (8B), LLaMA3.2 (3B), GPT-2, and Phi-4 (13B). For each LLM, five candidate solutions were generated, and the best result was selected for evaluation. Figure 6 presents the performance differences among these models, with evaluation metrics defined in Equations (6) - (9). It can be observed that LLaMA3.1 (8B) achieves the best dead-space control among all tested LLMs, while Phi-4 (13B) performs comparably to LLaMA3.2 (3B) and does not show a clear advantage. In contrast, GPT-2 exhibits the weakest performance. Given its superior balance between solution quality and training efficiency, LLaMA3.1 was selected as the primary model for subsequent comparisons against two variants of simulated annealing (SA).

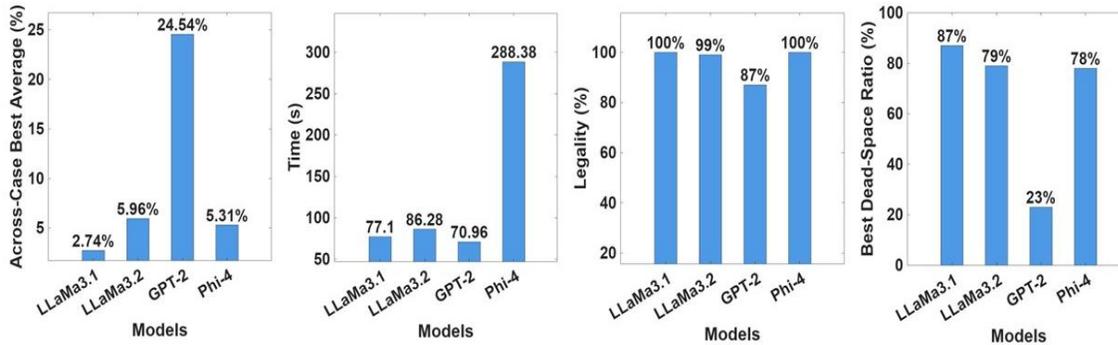

(a) Across-Case Best Average of Four LLMs  (b) Time of Four LLMs
(c) Legality of Four LLMs  (d) Best Dead-Space Ratio of Four LLMs

Figure 6. Comparison of Four Large Language Models

### 4.1 Dataset Generation Stage

#### 4.1.1 Across-Case Best Average:

As shown in Figure 7(a), LLaMA3.1 maintains a low dead-space ratio (approximately 2.74%–8.04%) across different module counts, clearly outperforming SA-Fast (approximately 9.95%–23.44%), which worsens rapidly with scale. SA-Quality achieves ratios comparable to or slightly lower than LLaMA3.1, but only at the cost of extremely high runtime. Overall, LLaMA3.1 balances solution quality and efficiency, underscoring the potential of data-driven approaches in 3D space planning.

#### 4.1.2 Time:

As shown in Figure 7(b), the inference time of LLaMA3.1 is comparable to SA-Fast (70–140 seconds) and grows linearly with the number of modules. In contrast, SA-Quality takes 800–1500 seconds, far exceeding both methods. This suggests that under practical time constraints, LLaMA3.1 can deliver good solution quality with much lower runtime, while SA-Quality may achieve lower dead space only at the cost of substantially reduced efficiency.

#### 4.1.3 Legality:

As shown in Figure 7(c), LLaMA3.1 maintains a legality rate between 96% and 100% across different module sizes, demonstrating that most generated sequences can be reconstructed into valid 3D slicing trees. In contrast, SA consistently achieves 100% legality as expected for heuristic methods. Although LLaMA3.1 occasionally produces invalid outputs, their proportion is very small and does not significantly affect overall performance, suggesting that the model has effectively learned stable slicing patterns.

#### 4.1.4 Best Dead-Space Ratio:

As illustrated in Figure 7(d), LLaMA3.1 shows a strong advantage in generating perfect solutions (0% dead space). For example, with 8 modules it achieves 87%, while SA-Fast reaches only 11% and SA-Quality 29%. Although the perfect-solution rate of LLaMA3.1 decreases as the number of modules grows—dropping to 57% at 16 modules—it consistently outperforms both SA variants by a large margin. This highlights the ability of LLaMA3.1 to deliver substantially more perfect layouts while maintaining efficiency.

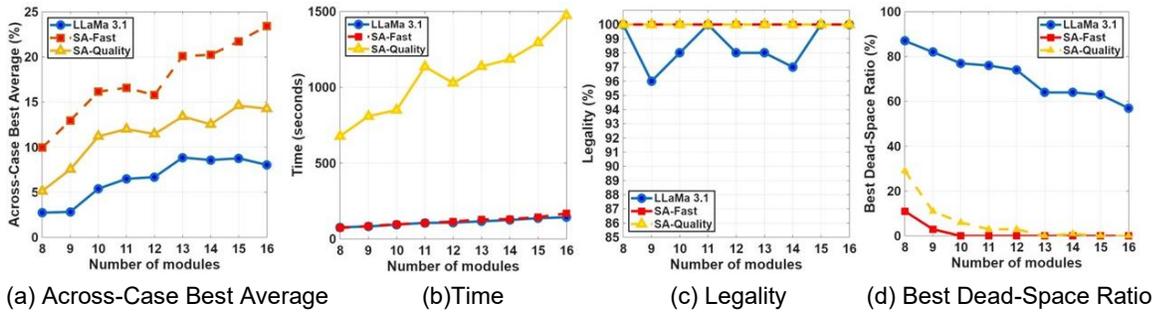

(a) Across-Case Best Average     (b)Time     (c) Legality     (d) Best Dead-Space Ratio

Figure 7. Comparison of Best Dead-Space Ratios between LLaMA3.1 and SA

Taken together, the results show that LLaMA3.1 achieves a strong balance between efficiency and quality: its dead-space ratio approaches that of SA-Quality, its perfect-solution rate clearly surpasses both SA variants, and its runtime remains comparable to SA-Fast. These findings highlight the potential of LLMs for practical applications, particularly when rapid generation of feasible space plans is required. While LLaMA3.1 still struggles to consistently deliver perfect solutions in higher-module cases, its promising performance indicates that with larger models and more training data, it could further close—or even surpass—the gap with traditional optimization methods.

### 4.2 Results on MCNC Benchmarks

To further evaluate the applicability of our approach to realistic design scenarios, we conducted experiments on MCNC-derived 3D benchmarks, including ami33, ami49, hp, xerox, and apte. These benchmarks are widely adopted in the EDA community and represent circuit layouts with diverse aspect ratios, making them closer to practical space planning tasks than synthetic datasets.

For ami33 and ami49, the large module counts make direct inference infeasible for current LLMs. As described in Section 3.1.4, we first partitioned these benchmarks into composite modules (11 for ami33 and five groups of sizes 10, 10, 10, 10, 9 for ami49). The bounding boxes of these composites were computed and then passed to the fine-tuned LLM for post-order slicing tree generation. The resulting layouts were subsequently refined with a one-dimensional compaction step. This pipeline enabled tractable evaluation while maintaining fidelity to the original benchmark characteristics.

The results are summarized in Table 1, which reports the legality rate and across-case best average dead-space ratio for each benchmark.

| Benchmark | Legality (%) | Across-Case Best Average (%) |
|---|---|---|
| apte | 92% | 31.30% |
| xerox | 98% | 51.71% |
| hp | 92% | 57.46% |
| ami33 | 88% | 75.02% |
| ami49 | 76% | 81.77% |

Table 1. Legality and Across-Case Best Average on MCNC Benchmarks

From Table 1, the proposed LLM-based framework achieves legality rates ranging from 76% to 98% across all MCNC benchmarks. The apte, xerox, and hp cases maintain legality above 90% with dead-space ratios between 31% and 57%, suggesting stable performance on medium-scale benchmarks. By contrast, for ami33 and ami49, legality decreases to 88% and 76%, and the dead-space ratios increase sharply to 75% and 82%, respectively. This trend highlights the inherent difficulty of scaling to dozens of modules with irregular aspect ratios.

Taken together, these results demonstrate that while the framework generalizes well to small- and medium-scale MCNC benchmarks, challenges remain for larger cases. The grouping and bounding-box preprocessing allowed the LLM to produce valid layouts, but the elevated dead-space ratios reveal scalability limitations. Addressing this gap—through larger training datasets, improved model architectures, or hybrid methods that combine LLM inference with heuristic refinement—will be critical for future extensions toward industrial-scale 3D IC designs.

## 5 CONCLUSION

### 5.1 Summary of Contributions

In this study, we proposed a large language model (LLM)-based framework for 3D IC space planning, leveraging post-order slicing tree representations to directly generate legal layouts with minimized dead space. By fine-tuning LLMs on large-scale synthetic and MCNC-derived datasets, the framework achieved strong performance in terms of solution quality, runtime efficiency, and stability. Among tested models, LLaMA3.1 (8B) demonstrated the best balance between training efficiency and layout quality, consistently achieving lower dead-space ratios, a higher proportion of zero-dead-space solutions, and comparable runtime to fast simulated annealing (SA-Fast).

Beyond synthetic datasets, we further validated the framework on MCNC-derived benchmarks, where it maintained high legality rates and robust generalization to realistic circuit footprints. Although larger and more irregular cases such as ami33 and ami49 remain challenging, the results confirm that the proposed approach can extend beyond artificial datasets and adapt to industry-standard test cases.

These findings highlight the potential of LLMs to serve as a data-driven alternative to traditional heuristic algorithms in electronic design automation (EDA). Furthermore, the framework shows strong adaptability to cross-domain applications such as logistics and 3D object placement. A lightweight compaction step was also incorporated to further reduce residual gaps without additional runtime overhead.

Overall, this work represents the first attempt to apply LLMs to 3D IC space planning, establishing a foundation for future research into scalable, multi-objective, and industry-ready applications.

## 6 LIMITATIONS AND FUTURE WORK

Although this study demonstrates the feasibility of leveraging large language models (LLMs) for 3D IC space planning, several limitations remain that open promising directions for future research:

### 6.1 Dataset scale and diversity:

The training corpus primarily consisted of 120,000 synthetic slicing-tree instances and 200,000 enlarged-cuboid instances, supplemented by MCNC-derived benchmarks. While this scale sufficed to demonstrate feasibility, it remains limited compared to industrial requirements, where hundreds of thousands of modules with diverse aspect ratios and heterogeneous constraints must be handled.

Future work should therefore expand dataset diversity to include industry-grade benchmarks, heterogeneous module distributions, and cases incorporating real design constraints. Validation on actual 3D IC layouts will be critical for demonstrating practical generalizability.

### 6.2 Scalability and module count:

The experiments in this study were conducted on problems with 8–16 modules and composite versions of MCNC benchmarks (ami33, ami49). These sizes provide a tractable setting for proof-of-concept, but industrial designs often involve hundreds or thousands of modules. Scaling LLMs to such problem sizes introduces challenges in terms of runtime efficiency, convergence stability, and legality of generated solutions. Future research should explore larger model architectures, improved training strategies, and hybrid approaches (e.g., combining LLM generation with heuristic refinement) to achieve scalability in industrial contexts.

### 6.3 Algorithmic comparison and benchmarking:

The evaluation in this study was limited to comparisons with simulated annealing (SA). While the results highlight advantages in runtime and dead-space reduction, a comprehensive understanding of LLM capabilities requires systematic benchmarking against recent approaches, including deep reinforcement learning (DRL), graph neural networks (GNNs), and diffusion-based generative models. Such broader comparisons will clarify the unique strengths and trade-offs of LLM-based planning relative to other machine learning paradigms.

### 6.4 Simplified optimization objectives:

This work focused primarily on minimizing dead space as the sole optimization criterion. However, practical 3D IC design requires balancing multiple objectives, including power delivery, thermal dissipation, interconnect wirelength, and TSV placement. Future research should extend the framework to multi-objective optimization, potentially through multi-task learning or reinforcement learning fine-tuning, to address real-world design trade-offs.

### 6.5 Resource and deployment constraints:

The current study was conducted under limited computational and manpower resources, using a single-server GPU environment and focusing on modest-scale benchmarks. Future extensions may involve distributed training, cloud-based deployment, and integration into EDA workflows. Incorporating digital twin platforms could further enable real-time simulation, monitoring, and optimization of IC layouts.

### 6.6 Resource and deployment constraints:

Beyond 3D IC design, the proposed framework has strong potential in logistics, cargo loading, and 3D object placement tasks where spatial efficiency is critical. Exploring these domains will not only broaden the applicability of the method but also provide additional insights into its generalization ability.

# Authors' background

| Your Name | Title* | Research Field | Personal website |
|---|---|---|---|
| **Hung-Ying Chu** | Undergraduate Student | EDA, Algorithm | N/A |
| **GUAN-WEI CHEN** | Undergraduate Student | EDA, Algorithm | N/A |
| **Shao-Yu Wei** | Undergraduate Student | EDA, Algorithm | N/A |
| **Yu-Cheng Lin** | Assistant Professor | EDA, Algorithm, AI music | https://www.cse.yzu.edu.tw/en/people/professor?name=Yu-Cheng%20Lin |